\documentclass{article}

\usepackage{microtype}
\usepackage{graphicx}
\usepackage{subcaption}
\usepackage{booktabs}
\usepackage{hyperref}
\usepackage{amsmath}
\usepackage{amssymb}
\usepackage{mathtools}
\usepackage[capitalize,noabbrev]{cleveref}
\usepackage{enumitem}
\usepackage[textsize=tiny]{todonotes}

\usepackage[accepted]{icml2026}
  
\newcommand{\topoloss}{\mathcal{L}_{\mathrm{topo}}}
\newcommand{\Mu}{M_u}
\newcommand{\aeff}{\alpha_{\mathrm{eff}}}

\icmltitlerunning{Topographic Training Concentrates Causal Circuits}

\begin{document}

\twocolumn[
  \icmltitle{Topographic Training Concentrates Causal Circuits\\
    Without Improving Neuron Monosemanticity}

  \begin{icmlauthorlist}
    \icmlauthor{Gautam Ranka}{inst}
    \icmlauthor{Shubham Pandere*}{inst}
    \icmlauthor{Aiden Ross D'souza*}{inst}
  \end{icmlauthorlist}
  \icmlaffiliation{inst}{IvLabs, VNIT Nagpur, India}
  \icmlcorrespondingauthor{Gautam Ranka}{gautam.ranka@ivlabs.in}
  \icmlcorrespondingauthor{Shubham Pandere}{shubham.pandere@ivlabs.in}
  \icmlcorrespondingauthor{Aiden Ross D'souza}{aiden.dsouza@ivlabs.in}

  \icmlkeywords{mechanistic interpretability, superposition, topographic
    neural networks, sparse autoencoders, activation patching}

  \vskip 0.3in
]

\printAffiliationsAndNotice{}

\begin{abstract}
Mechanistic interpretability of vision transformers seeks to decompose model computation into human-readable 
  units, but learned representations entangle many concepts in each neuron. Feature superposition is
  widely treated as the central obstacle to this decomposition, yet most mitigations (sparse autoencoders, dictionary  
  learning) are post-hoc and leave the underlying network unchanged. We ask whether a spatial-locality
  training loss (TopoLoss) can act as a lightweight, training-time prior that improves interpretability
  of standard mech-interp tools. Training ViT on ImageNet-100 across multiple TopoLoss weights $\alpha$,
  we measure causal sufficiency of topographic clusters via activation patching and feature geometry via sparse
  autoencoders fit to the same residual stream. At $\alpha=1.0$, topographic clusters are
  2.79$\times$ more causally sufficient than random unit sets of the same size, with the effect increasing monotonically in $\alpha$. SAE L0
  sparsity decreases by 11\% and dead-feature fraction rises 19-fold, yet standard neuron-level
  monosemanticity scores are unchanged, indicating that topographic pressure acts at circuit level, concentrating causal mass into spatially local structures without disentangling individual neurons.
  This dissociation suggests current neuron-level monosemanticity metrics
  are insensitive to a class of real interpretability gains, and positions cheap architectural priors as a
  viable training-time complement to post-hoc tooling.
\end{abstract}

\section{Introduction}
\label{sec:intro}

The superposition hypothesis~\citep{elhage2022superposition} posits that
neural networks represent more features than they have neurons, encoding
multiple unrelated concepts in each unit through approximate orthogonality
in high-dimensional space.
Superposition makes circuit analysis and in general, mechanistic analysis hard: patching a single neuron
contaminates unrelated features, and sparse autoencoders (SAEs) must
expend capacity to decompose polysemantic units into monosemantic
directions~\citep{bricken2023monosemanticity,cunningham2023saes}.
Current solutions are almost entirely \emph{post-hoc} where they learn to
decode a fixed representation rather than encouraging the model to form
a cleaner one.

A complementary strategy is to impose an inductive bias during training
that couples nearby neurons, hoping spatial locality produces a form of
\emph{functional organisation} analogous to cortical maps in biological
vision~\citep{khosla2025toponets}.
\citeauthor{khosla2025toponets} recently showed that a Laplacian weight-smoothness
penalty (TopoLoss) applied to vision and language transformers improves
brain-similarity scores without sacrificing accuracy.
The key open question is whether this reorganisation also improves
\emph{mechanistic interpretability metrics}.

We conduct a systematic audit of TopoLoss across three interpretability axes:
\begin{itemize}[noitemsep,topsep=1pt]
  \item \textbf{H1 (monosemanticity):} Do individual neurons become more
    selective? Measured via entropy-based selectivity $\Mu$.
  \item \textbf{H2 (superposition):} Does the SAE-measured feature complexity
    decrease? Measured via L0 norm and dead-feature fraction.
  \item \textbf{H3 (causal purity):} Are topographic clusters more causally
    sufficient for class prediction than random unit sets?
    Measured via activation-patching logit-delta ratios.
\end{itemize}

\noindent
Our main findings are: H3 is strongly supported (p$<$0.0001, $d=0.95$);
H2 is supported at high $\alpha$ with a confound of accuracy;
H1 is null.
We further observe that H3 effects are more modest in a smaller
4-layer architecture (TinyViT), consistent with a scale or depth
requirement for circuit formation, though additional experiments
are needed to isolate this factor.

\section{Background and Method}
\label{sec:method}

\begin{figure*}[t]
  \vskip 0.1in
  \begin{center}
    \includegraphics[width=0.96\textwidth]{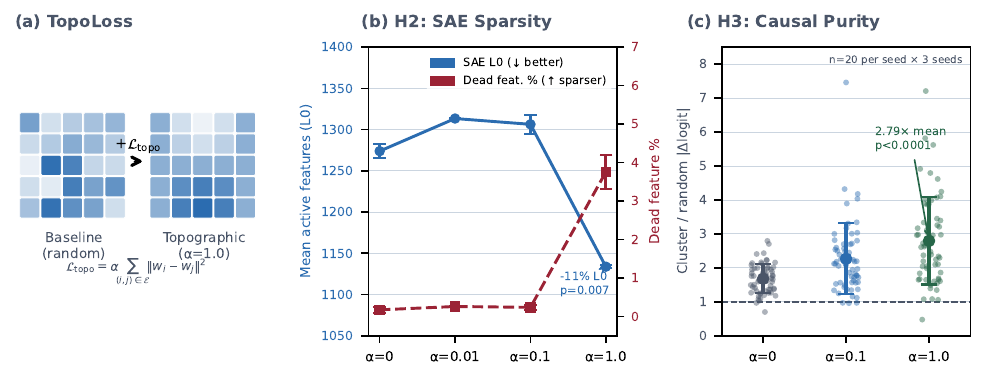}
  \end{center}
  \vskip -0.1in
  \caption{%
    \textbf{Overview.}
    \emph{(a)} TopoLoss (Eq.~\ref{eq:topoloss}) penalises weight differences
    between spatially adjacent neurons, pushing functionally similar units
    to co-localise on the topographic grid.
    \emph{(b)} H2: SAE L0 (mean active features, blue) decreases monotonically with $\alpha$, while dead-feature fraction (red, dashed) rises 19-fold at $\alpha=1.0$ (y-axis truncated; baseline L0 = 1273.9).
    \emph{(c)}  H3: per-class causal patching ratios($n=60$, 20 classes $\times$ 3 seeds). Points show individual class$\times$seed pairs; filled markers show mean $\pm$ SD\@. Topographic clusters at $\alpha=1.0$ are $2.79\times$ more causally sufficient than random unit sets ($p<0.0001$, $d=0.95$). The dashed line marks the  random-cluster baseline (ratio $=1$).
  }
  \label{fig:overview}
\end{figure*}

\paragraph{TopoLoss.}
Following \citet{khosla2025toponets}, we add a weight-smoothness penalty
to the standard cross-entropy loss:
\begin{equation}
  \mathcal{L} = \mathcal{L}_{\mathrm{CE}} + \topoloss, \qquad
  \topoloss = \alpha \!\!\sum_{(i,j)\in\mathcal{E}} \|w_i - w_j\|^2,
  \label{eq:topoloss}
\end{equation}
where $\mathcal{E}$ is the set of adjacent unit pairs on a fixed 2D
topographic grid and $\alpha\geq 0$ controls pressure strength.
In our ViT-S implementation, TopoLoss is applied to the attention-projection
weights (\texttt{attn.proj}) of all 12 blocks; units are arranged on a
$\sqrt{d}\times\sqrt{d}$ grid matching the head dimension.

\paragraph{Models and training.}
We train ViT-S/16~\citep{dosovitskiy2021vit} on
ImageNet-100~\citep{deng2009imagenet} (126,688 train / 5,000 val,
100 classes) for $\alpha \in \{0.0, 0.1, 1.0\}$, with three seeds
$\{42, 123, 456\}$ each, total of nine checkpoints.
A broader $\alpha$-sweep ($\alpha \in \{0, 0.01, 0.1, 1.0\}$, seed 42)
provides dose-response curves.
We additionally train a lightweight TinyViT (4 layers, $d=128$, 4 heads)
with the same protocol as a preliminary scale probe.
All runs use AdamW, cosine schedule, and standard augmentation;
full hyperparameters are in \cref{app:training}.

\paragraph{Evaluation metrics.}

\emph{H1 : Monosemanticity} $\Mu$ is the normalised negative entropy of
a neuron's class-activation distribution over 5,000 validation images,
$\Mu = (H_{\max} - H)/H_{\max}$, so $\Mu=1$ is perfectly selective.
We also train an SAE (TopK, $k=32$, 2048 features) on layer-6 residuals
and measure per-feature selectivity. 

\emph{H2 : SAE L0 sparsity.}
We train one SAE per checkpoint at layer 6 (\texttt{attn.proj} output)
and report the mean number of active features per token (L0) and the
fraction of dead features (active on $<$0.1\% of tokens).
Lower L0 and higher dead-feature fraction indicate sparser, more
specialised representations. The choice of layer 6 was motivated by it lying in the middle of the effect band (Appendix \ref{app:layerwise})

\emph{H3 : Causal purity.}
For each of 20 randomly selected ImageNet-100 classes and each of the
3 seeds, we identify the top-$k=16$ neurons most activated by that class,
compute the mean absolute logit delta when patching this cluster
from a source (target-class image) into a target (other-class image),
and normalise by the delta from patching a random size-$k$ set.
Ratios $>1$ mean topographic clusters carry disproportionate causal weight.
The top-$k$ selection rule is by activation magnitude and does not enforce spatial contiguity;
we verify post-hoc (Table~\ref{tab:spatial}) that these sets are nonetheless spatially co-localised
under TopoLoss, justifying the cluster framing.

\section{Results}
\label{sec:results}

\subsection{H3: Topographic Clusters Are Causally Sufficient}
\label{sec:h3}

\begin{figure*}[t]
  \vskip 0.1in
  \begin{center}
    \includegraphics[width=0.9\textwidth]{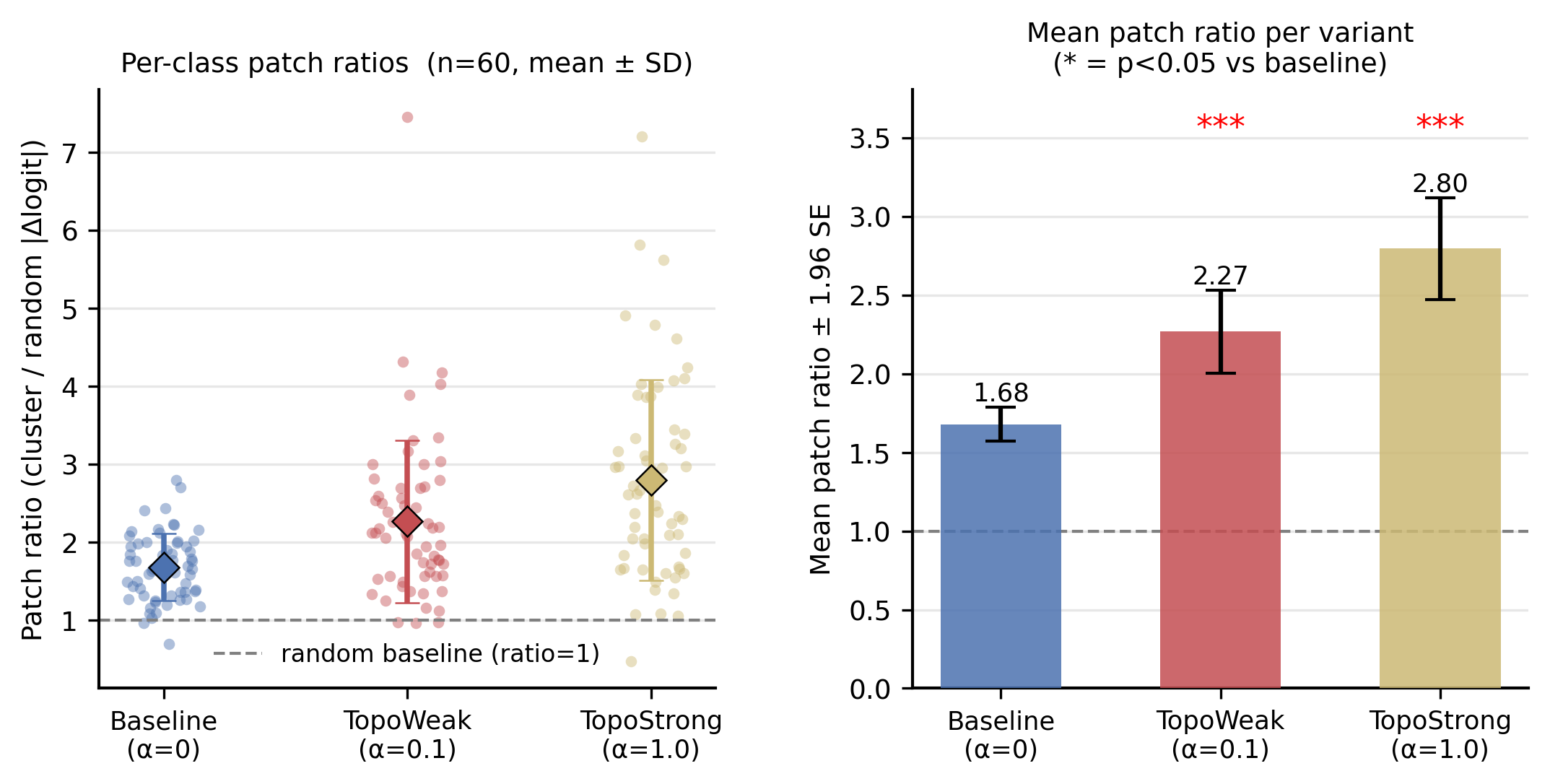}
  \end{center}
  \vskip -0.1in
  \caption{%
    \textbf{H3: Causal patching ratios across 20 classes $\times$ 3 seeds
    ($n=60$).}
    Each point is the mean cluster/random logit-delta ratio for one
    class$\times$seed combination.
    The ratio at $\alpha=1.0$ ($2.79\pm1.24$) is significantly above
    baseline ($1.68\pm0.42$), and above $\alpha=0.1$ ($2.27\pm0.93$).
    Both comparisons: $p<0.0001$, Cohen's $d\geq0.61$.
  }
  \label{fig:h3}
\end{figure*}

Table~\ref{tab:h3_main} and Figure~\ref{fig:h3} show the primary result.
Topographic clusters at $\alpha=1.0$ produce a mean patching ratio of
$2.79\pm1.24$, versus $1.68\pm0.42$ for the baseline leading to a 66\% increase.
The effect is monotone in $\alpha$ ($1.68\to2.27\to2.79$) and robust
across all 20 classes and all 3 seeds independently
(per-seed $\alpha=1.0$ ratios: 2.53, 2.89, 2.97).
A paired $t$-test on seed-level means yields $p=0.082$ ($n=3$), which
is underpowered; the multi-class expansion to $n=60$ gives $p<0.0001$,
$d=0.95$ (\cref{tab:h3_main}).

\begin{table}[t]
  \caption{%
    \textbf{H3 patching ratios} (cluster / random $|\Delta\text{logit}|$),
    multi-class $n=60$.
  }
  \label{tab:h3_main}
  \vskip 0.05in
  \begin{center}
    \begin{small}
      \begin{tabular}{lcccc}
        \toprule
        $\alpha$ & Mean ratio & $\Delta$ vs base & $p$ & $d$ \\
        \midrule
        0.0 & $1.68\pm0.42$ & — & — & — \\
        0.1 & $2.27\pm0.93$ & $+0.59$ & $<0.0001$ & 0.61 \\
        1.0 & $2.79\pm1.24$ & $+1.12$ & $<0.0001$ & 0.95 \\
        \bottomrule
      \end{tabular}
    \end{small}
  \end{center}
  \vskip -0.1in
\end{table}

\paragraph{Spatial coherence of top-k unit sets.}
To verify that the H3 selection rule picks out spatially co-localised units under TopoLoss
(and not merely high-activation units distributed arbitrarily across the grid), we compute
the mean pairwise Euclidean grid distance among the top-$k=16$ class units on the
$16\times24$ topographic sheet. Lower values indicate tighter spatial concentration;
the random-baseline expected distance for a size-16 subset is 10.618. Results across
20 classes $\times$ 3 seeds ($n=60$) are shown in Table~\ref{tab:spatial}.
Top-k units at $\alpha=1.0$ are 6.1\% more spatially concentrated than baseline
($p=0.0005$, $d=0.66$); the $\alpha=0.1$ effect is not significant, mirroring
the H3 threshold pattern.

\begin{table}[t]
  \caption{%
    \textbf{Spatial coherence of top-k=16 class units}
    (mean pairwise grid distance on the $16\times24$ sheet;
    lower = more spatially concentrated; random baseline = 10.618).
    $n=60$ (20 classes $\times$ 3 seeds).
  }
  \label{tab:spatial}
  \vskip 0.05in
  \begin{center}
    \begin{small}
      \begin{tabular}{lcccc}
        \toprule
        $\alpha$ & Mean dist & $\Delta$\% vs base & $p$ & $d$ \\
        \midrule
        0.0 & $10.599\pm0.892$ & ---            & ---    & --- \\
        0.1 & $10.525\pm0.934$ & $-0.7\%$       & 0.661 (n.s.) & 0.08 \\
        1.0 & $\mathbf{9.948\pm1.063}$ & $\mathbf{-6.1\%}$ & $\mathbf{0.0005}$ & $\mathbf{0.66}$ \\
        \bottomrule
      \end{tabular}
    \end{small}
  \end{center}
  \vskip -0.1in
\end{table}

\subsection{H2: Reduced Feature Superposition}
\label{sec:h2}

At $\alpha=1.0$, SAE L0 decreases by $11\%$ relative to baseline
($1133.8\pm3.2$ vs $1273.9\pm15.0$; $p=0.007$, $d=7.06$) and the
dead-feature fraction rises 19-fold ($3.8\pm0.8\%$ vs $0.2\pm0.1\%$).
The $\alpha=0.1$ model shows neither effect reliably ($p=0.054$ for L0,
$p=0.580$ for dead fraction), confirming a threshold between $\alpha=0.1$
and $\alpha=1.0$.
The L0 reduction is distributed across depth (layers 3--10, peak at
layer 10; see Appendix~\ref{app:layerwise}), not localised to a single block.

\paragraph{Confound.}
We treat the L0 reduction as suggestive but not confound-free.
The $\alpha=1.0$ model suffers a $-1.6$ pp accuracy cost
($61.2\% \to 59.6\%$), and an OLS regression of L0 on validation
accuracy across all 9 ViT-S checkpoints yields $R^2=0.72$ ($p=0.004$),
indicating accuracy explains substantial L0 variance.
Two analyses suggest the effect is not entirely accuracy-driven.
First, the PCA $d_{95}$ dissociation at $\alpha=0.1$, where SAE L0
\emph{increases} ($+32.5$) while $d_{95}$ \emph{decreases} ($-8.3$),
shows SAE L0 captures structure beyond raw effective dimensionality.
Second, two of three $\alpha=1.0$ seeds lie below the accuracy-L0
regression line (sparser than accuracy alone predicts).
An accuracy-matched ablation would directly resolve the confound
and is left for future work.

\subsection{H1: No Change in Neuron Monosemanticity}
\label{sec:h1}

Entropy-based $\Mu$ is flat across $\alpha$
($0.234\pm0.002$ at $\alpha=0$ vs $0.236\pm0.002$ at $\alpha=1.0$;
$\Delta=0.0019$, $n.s.$).
SAE-feature-level $\Mu$ is also null.
Full tables are in \cref{app:h1}.
The H1 null alongside the H3 positive indicates that topographic pressure
reorganises the circuit-level feature geometry without altering
the per-neuron class-selectivity distribution.

\subsection{Scale Dependence: TinyViT}
\label{sec:tinyvit}

As a preliminary probe, we trained TinyViT (4-layer, $d=128$) with the
same protocol.
The H2 trend replicates: L0 decreases monotonically with $\alpha$ ($-2.1$\%
at $\alpha=1.0$), consistent with the weight-norm scaling rule
$\aeff = \alpha \cdot (\mathrm{rms}_{\mathrm{ref}}/\mathrm{rms}_{\mathrm{tgt}})^2$
(see \cref{app:scaling}).
H3, however, saturates: the maximum ratio across all $\alpha$ is $1.42\times$
($\alpha=0.1$), far below the ViT-S result of $2.79\times$.
We caution that TinyViT and ViT-S differ simultaneously in depth, width,
accuracy, and training budget; with a single seed per $\alpha$ we cannot
isolate the responsible factor.
Depth is one plausible hypothesis as a 4-layer network may lack sufficient
representational stages for topographic pressure to form coherent causal
circuits, but this requires a controlled depth-sweep to confirm.
We treat the TinyViT result as a useful negative: the simple weight-norm
calibration rule predicts H2 transfer (which we observe) but not H3
transfer (which we do not), suggesting H3 depends on architectural factors
beyond effective topographic pressure.

\section{Discussion}
\label{sec:discussion}

\paragraph{Why H3 but not H1?}
Monosemanticity ($\Mu$) measures the distribution of a single neuron over
classes, a marginal statistic insensitive to the \emph{joint} arrangement
of multiple neurons.
Activation patching measures whether a specific set of neurons is jointly
sufficient for a prediction, a multivariate, \emph{circuit-level} probe.
Our results suggest that TopoLoss reorganises neurons into functionally
coherent spatial clusters (H3) without changing how selectivity is spread
across individual neurons (H1).
This is consistent with \citet{elhage2022superposition}: superposition
redistributes information across neurons, so spatial co-location (not
per-neuron selectivity) is the relevant unit of circuit purity.

\paragraph{TopoLoss as a training-time interpretability prior.}
Standard mech-interp interventions are post-hoc: SAEs learn to decode
a representation after training; patching reveals circuits but does not
change them.
Our results show that a $<$0.1k-parameter regularisation term, added to
the training loss, systematically shifts circuit geometry in a way
measurable by existing tools.
The effect is modest in cost ($-1.6$ pp accuracy at $\alpha=1.0$) and
strong in signal ($d=0.95$ for H3).

\paragraph{Related work.}
\citet{khosla2025toponets} demonstrated that TopoLoss improves brain
similarity without accuracy loss; we evaluate it on mechanistic
interpretability axes for the first time.
\citet{bricken2023monosemanticity} and \citet{cunningham2023saes} use SAEs
as diagnostic tools; we show that TopoLoss \emph{shifts} the SAE's diagnosis.
\citet{conmy2023acdc} use activation patching to localise circuits within
a single model; we use the same protocol to \emph{compare} models trained
with different priors.
Topographic constraints in deep networks have a longer lineage
\citep{lee2020,doshi2023,margalit2024}; we focus on TopoLoss as a
recent, lightweight instantiation amenable to standard mech-interp tooling.
Work on topographic self-organisation in networks
(\citealp{zeiler2014visualizing}; \citealp{olah2020zoom}) has noted
functional clustering informally; we quantify it with controlled patching
experiments.

\paragraph{Limitations.}
\begin{itemize} [noitemsep,topsep=1pt]
    \item Single dataset (ImageNet-100).
    \item H2 has a partial accuracy confound, addressed but not fully
resolved.
    \item Explored only on discriminatory vision transformer models.
    \item The $\Mu$ metric uses ImageNet-100 classes as the concept basis,
which may be too coarse to detect monosemanticity in features that cut
across classes (e.g.\ textures, parts). The 25k-image SAE-feature $\Mu$
evaluation (Appendix~\ref{app:h1}) addresses the sample-size concern but
inherits the same class-basis limitation.
\end{itemize}

\paragraph{Future work.}
\begin{itemize}[noitemsep,topsep=1pt]
    \item A depth-controlled sweep (TinyViT $\to$ ViT-S $\to$ ViT-B at fixed width)
would test the scale hypothesis.
    \item Extension to other conventional vision architectures like CNNs and ResNets.
    \item An embedding-based monosemanticity score that sidesteps the class basis \citet{pach2025sae}, would provide an independent test of whether the null holds, applied either to raw neurons or to the SAE-recovered features.
    \item Brain-similarity alignment (NSD fMRI), SOM post-hoc control, and language-model replication are natural extensions.
\end{itemize}

\section*{Impact Statement}

This paper presents work whose goal is to advance the field of mechanistic
interpretability by studying training-time interventions that reduce feature
superposition.
Improved interpretability tools may help identify and mitigate harmful
capabilities in AI systems.
There are no direct negative societal consequences we foresee from this
methodological contribution.

\section{Code}
The code repository can be accessed at \url{https://github.com/IvLabs/toposae}.

\bibliography{references}
\bibliographystyle{icml2026}

\newpage
\appendix
\onecolumn

\section{Training Details}
\label{app:training}

All models trained with AdamW ($\beta_1=0.9$, $\beta_2=0.999$,
$\epsilon=10^{-8}$), cosine learning-rate schedule, weight decay 0.05,
warm-up 10 epochs.
ViT-S: batch 256, lr $5\times10^{-4}$, up to 100 epochs with early stopping
(patience 15).
TinyViT: same schedule, batch 128, lr $3\times10^{-4}$, 50 epochs.
SAE training: TopK sparse autoencoder, $k=32$, 2048 features, Adam,
lr $10^{-3}$, 50 epochs.
All experiments run on a single NVIDIA RTX 3050 (4 GB) for TinyViT and
analysis; ViT-S checkpoints trained on a remote A100.

\begin{table}[h]
  \caption{Training results summary (ViT-S/16, multi-seed).}
  \label{tab:training}
  \begin{center}
    \begin{tabular}{llcc}
      \toprule
      $\alpha$ & Label & Val acc (mean$\pm$std) & Epochs \\
      \midrule
      0.0 & Baseline   & $0.612\pm0.005$ & $\sim87$ \\
      0.1 & TopoWeak   & $0.614\pm0.002$ & $\sim67$ \\
      1.0 & TopoStrong & $0.596\pm0.005$ & $\sim79$ \\
      \bottomrule
    \end{tabular}
  \end{center}
\end{table}

\section{H1: Full Monosemanticity Tables}
\label{app:h1}

\begin{table}[h]
  \caption{Neuron-level entropy monosemanticity $\Mu$ (ViT-S/16,
    multi-seed; higher = more selective).}
  \label{tab:h1}
  \begin{center}
    \begin{tabular}{lccc}
      \toprule
      $\alpha$ & Mean $\Mu$ & Frac.\ $>0.5$ \\
      \midrule
      0.0 & $0.2340\pm0.0019$ & 0.26\% \\
      0.1 & $0.2348\pm0.0025$ & 0.52\% \\
      1.0 & $0.2359\pm0.0016$ & 0.78\% \\
      \bottomrule
    \end{tabular}
  \end{center}
\end{table}

No $\alpha$-level comparison of neuron-level $\Mu$ is statistically significant.

\paragraph{SAE-feature $\Mu$ at 25k images (EXP\_012b).}
The original SAE-feature $\Mu$ evaluation (EXP\_012, 5k validation images)
was undersampled. We re-ran the analysis on 25{,}000 stratified training
images (250 per class, expansion factor 4, 1536 features, $L_1=0.005$,
50 epochs); see Table~\ref{tab:h1_sae25k}. Mean per-feature $\Mu$ remains
flat across $\alpha$ (0.021, 0.021, 0.026), and the fraction of features
with $\Mu > 0.5$ is zero throughout. The H1 null therefore holds in both
the neuron basis and the SAE-feature basis at the larger sample size,
strengthening the H1/H3 dissociation.

\begin{table}[h]
  \caption{SAE-feature $\Mu$ on 25k stratified train images
    (expansion factor 4, 1536 features, $n=3$ seeds per $\alpha$).}
  \label{tab:h1_sae25k}
  \begin{center}
    \begin{small}
      \begin{tabular}{lcccc}
        \toprule
        $\alpha$ & SAE L0 & Mean $\Mu$ & Frac.\ $>0.3$ & Frac.\ $>0.5$ \\
        \midrule
        0.0 & 1304 & 0.021 & 0.000 & 0.000 \\
        0.1 & 1316 & 0.021 & 0.000 & 0.000 \\
        1.0 & 1189 & 0.026 & 0.000 & 0.000 \\
        \bottomrule
      \end{tabular}
    \end{small}
  \end{center}
\end{table}

\section{H2: Confound Analyses}
\label{app:h2}

\paragraph{EXP\_010 — Accuracy-L0 OLS.}
OLS regression of SAE L0 on validation accuracy across all 9 ViT-S
checkpoints: $R^2=0.724$, $p=0.004$.
Residuals for the three $\alpha=1.0$ seeds: $-54$, $-37$, $+21$
(two below the regression line — sparser than accuracy alone predicts,
one above).
The pattern is suggestive but not conclusive across all seeds.

\paragraph{EXP\_011 — PCA dimensionality control.}

\begin{table}[h]
  \caption{PCA 95\%-variance dimension ($d_{95}$) vs SAE L0 (ViT-S/16).}
  \label{tab:pca}
  \begin{center}
    \begin{tabular}{lcccc}
      \toprule
      $\alpha$ & PCA $d_{95}$ & $\Delta d_{95}$ & SAE L0 & $\Delta$L0 \\
      \midrule
      0.0 & $166.7\pm3.3$ & — & 1273.9 & — \\
      0.1 & $158.3\pm3.8$ & $-8.3$ & 1306.4 & $+32.5$ \\
      1.0 & $142.3\pm3.4$ & $-24.3$ & 1133.8 & $-140.1$ \\
      \bottomrule
    \end{tabular}
  \end{center}
\end{table}

The dissociation at $\alpha=0.1$ (PCA $d_{95}$ decreases while L0
\emph{increases}) shows that SAE L0 captures representational geometry
beyond effective dimensionality.

\section{H3: Per-Seed Robustness}
\label{app:h3_perseed}

\begin{table}[h]
  \caption{Per-seed H3 ratios at $\alpha=1.0$ (single-class, seed-level).}
  \label{tab:h3_perseed}
  \begin{center}
    \begin{tabular}{lccc}
      \toprule
      Seed & Cluster $|\Delta\text{logit}|$ & Random & Ratio \\
      \midrule
      42  & 1.282 & 0.340 & 3.77$\times$ \\
      123 & 0.911 & 0.386 & 2.36$\times$ \\
      456 & 0.892 & 0.401 & 2.22$\times$ \\
      \bottomrule
    \end{tabular}
  \end{center}
\end{table}

All three seeds individually exceed $2\times$; the effect is not driven
by a single outlier seed.

\section{Layer-Wise SAE Analysis}
\label{app:layerwise}

\begin{figure}[h]
  \centering
  \includegraphics[width=0.9\columnwidth]{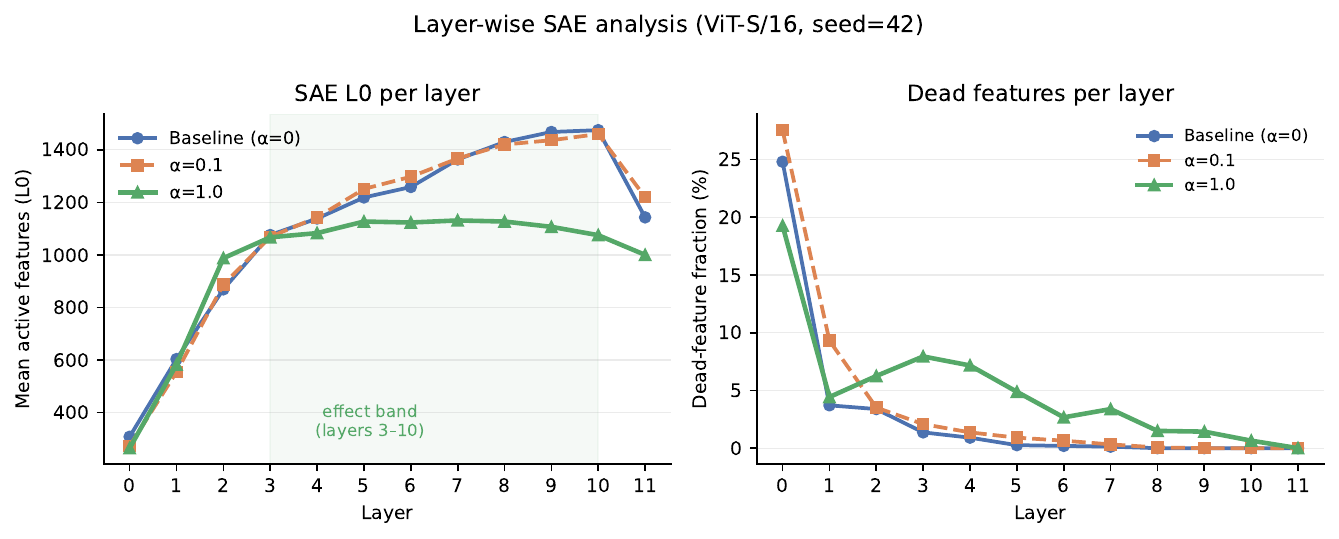}
  \caption{Layer-wise SAE L0 (seed 42). The $\alpha=1.0$ curve diverges
    from baseline at layer~3 and remains below through layer~10;
    $\alpha=0.1$ tracks baseline. Layer~6 (used for main-text H2/H3
    analyses) lies in the middle of the effect band.}
  \label{fig:layerwise}
\end{figure}

H2 effects are distributed across depth rather than concentrated in a
single layer (Figure~\ref{fig:layerwise}). The L0 gap between $\alpha=1.0$
and baseline first appears at layer~3 and widens through layers~5--10
(peak gap at layer~10). The $\alpha=0.1$ curve tracks baseline throughout.
The choice of layer~6 for the main-text H2/H3 analyses sits in the middle
of the effect band and is therefore representative.

\section{Alpha-Scaling Rule for H2}
\label{app:scaling}

TopoLoss gradient magnitude scales as $\nabla_w \topoloss \propto \alpha \cdot w$,
so the effective topographic pressure on weight magnitudes scales as
$\aeff \propto \alpha \cdot \mathrm{rms}(w)^2$.
Calibrating to a reference model (ViT-S, $\alpha=1.0$):
\[
  \aeff = \alpha \cdot \left(\frac{\mathrm{rms}_{\mathrm{ref}}}{\mathrm{rms}_{\mathrm{tgt}}}\right)^2.
\]

\begin{table}[h]
  \caption{Weight-norm scaling and effective $\alpha$.}
  \label{tab:scaling}
  \begin{center}
    \begin{tabular}{llcc}
      \toprule
      Model & $\alpha$ & attn.proj RMS & $\aeff$ \\
      \midrule
      TinyViT & 0.0 & 0.0695 & — \\
      TinyViT & 1.0 & 0.2346 & 2.6 (explosion) \\
      ViT-S   & 0.0 & 0.0432 & — \\
      ViT-S   & 1.0 & 0.0548 & 1.12 (stable) \\
      \bottomrule
    \end{tabular}
  \end{center}
\end{table}

At $\alpha=1.0$, TinyViT undergoes weight-norm explosion (RMS grows
$3.4\times$ from baseline) while ViT-S remains stable ($1.27\times$).
Using the calibration rule, TinyViT at $\alpha=0.38$ should match
ViT-S at $\alpha=1.0$ in effective pressure.
We ran this configuration and found H2 (L0) behaved as predicted
(L0 intermediate between $\alpha=0.1$ and $\alpha=1.0$) but H3 did not
improve over $\alpha=0.1$ ($1.11\times$ vs $1.42\times$).
This falsification suggests H3 depends on factors beyond pressure
calibration — model scale, depth, or training budget are candidate
explanations requiring a controlled sweep to isolate.

\section{TinyViT Full Results}
\label{app:tinyvit}

\begin{table}[h]
  \caption{TinyViT (4L/128D) results, seed=42, 50 epochs.}
  \label{tab:tinyvit}
  \begin{center}
    \begin{small}
      \begin{tabular}{lcccccc}
        \toprule
        $\alpha$ & Val acc & $\Mu$ & SAE L0 & Dead\% & H3 ratio \\
        \midrule
        0.0  & 0.472 & 0.253 & 444.9 & 2.0 & 1.19$\times$ \\
        0.1  & 0.462 & 0.260 & 443.7 & 2.5 & \textbf{1.42}$\times$ \\
        0.38 & 0.470 & 0.262 & 437.5 & 1.4 & 1.11$\times$ \\
        1.0  & 0.457 & 0.249 & 435.5 & 2.9 & 1.12$\times$ \\
        \bottomrule
      \end{tabular}
    \end{small}
  \end{center}
\end{table}

H3 peak is at $\alpha=0.1$, not $\alpha=1.0$ as in ViT-S.
Single-seed; interpret with caution.

\end{document}